\documentclass[letterpaper, 10 pt, conference]{ieeeconf}  

\IEEEoverridecommandlockouts                              

\usepackage{cite}
\usepackage{amsmath,amssymb,amsfonts}
\usepackage{algorithmic}
\usepackage{graphicx}
\usepackage{textcomp}
\usepackage{xcolor}
\usepackage{comment}
\title{\LARGE \bf
Evaluation of Performance-Trust vs Moral-Trust Violation in 3D Environment}

\author{Maitry Ronakbhai Trivedi$^{1}$  Zahra Rezaei Khavas $^{2}$ and Paul Robinette$^{3}$%
\thanks{$^{1}$Maitry Ronakbhai Trivedi is with the Department of Computer Science, University of Massachusetts Lowell, Lowell, MA, USA
        {\tt\small MaitryRonakbhai\_Trivedi@student.uml.edu}}%
\thanks{Zahra Rezaei Khavasr$^{2}$ and Paul Robinette$^{3}$ are with The Department of Electrical and Computer Engineering, University of Massachusetts Lowell, Lowell, MA, USA
        {\tt\small zahra\_rezaeikhavas@student.uml.edu} , {\tt\small paul\_robinette@uml.edu}}%
}

\begin{document}

\maketitle
\thispagestyle{empty}
\pagestyle{empty}

\begin{abstract}
Human-Robot Interaction, in which a robot with some level of autonomy interacts with a human to achieve a specific goal has seen much recent progress. With the introduction of autonomous robots and the possibility of widespread use of those in near future, it is critical that humans understand the robot's intention while interacting with them as this will foster the development of human-robot trust. The new conceptualization of trust which had been introduced by researchers in recent years considers trust in Human-Robot Interaction to be a multidimensional nature. Two main aspects which are attributed to trust are performance trust and moral trust. We aim to design an experiment to investigate the consequences of performance-trust violation and moral-trust violation in a search and rescue scenario. We want to see if two similar robot failures, one caused by a performance-trust violation and the other by a moral-trust violation have distinct effects on human trust. In addition to this, we plan to develop an interface that allows us to investigate whether altering the interface's modality from grid-world scenario (2D environment) to realistic simulation (3D environment) affects human perception of the task and the effects of the robot's failure on human trust.
\end{abstract}

\section{INTRODUCTION}

In recent years. researchers from the Human-Robot Interaction (HRI) field are encouraging the implementation of autonomous robots over teleoperated robots. The numerous applications include assisting humans by providing mechanical tools\cite{ardon2021affordance} \cite{ortenzi2022robot}, educating children \cite{westlund2016lessons}, replacing or assisting humans in rescue tasks \cite{murphy2005up} \cite{park2008measuring}, and so on. In this case, it is critical to determine how much autonomy the robot should have to maximize efficiency and minimize causalities. According to the definition by \cite{beer2014toward}, the Level of Autonomy (LoA) is ``The extent to which a robot can sense its environment, plan based on that environment, and act upon that environment with the intent of reaching some task-specific goal without external control''. They have proposed a guideline for determining the LoA based on the task criticality, task accountability, and environmental complexity. The majority of automation research has focused on the relationship between automation reliability and operator usage, with the intervening element of trust frequently being overlooked \cite{lewis2018role}. 

\par
As the autonomous agents becomes ubiquitous, trust in HRI becomes crucial in order to build a successful eco-system. Trust is a measure by which humans can assess the degree to which a robot is involved in a task. Trust in HRI is defined as a multidimensional entity that includes one aspect similar to trust in Human-Automation Interaction (HAI) and one aspect similar to trust in Human-Human Interaction (HHI) \cite{ullman2019measuring}. The human-automation trust places considerable emphasis on the autonomous system's performance. In other words, it determines how capable and efficient the system is. On the other hand, human-human trust focuses on whether humans exhibit moral behavior among each other by not leveraging others' vulnerabilities.

Our research intends to investigate whether two similar robot failures of equal magnitude affect human trust differently if one is influenced by a moral-trust violation and the other by a performance-trust violation. People react differently to the robot's performance and moral-trust violation, which can be seen in our prior work on a 2D grid-world game described in \ref{method2d}. Our preliminary results from the 2D game indicates that the moral trust violation has a significant effect on human-robot trust. Since the grid-world game is not as realistic as real world simulation scenario, it raises an important question of how human trust is influenced. Therefore, we aim to investigate whether changing the mode of interaction from grid-world (2D environment) to real world 3D simulation would influence human trust. The main purpose of this research is that we can quickly and simply blueprint various experiment designs in a 2D game before moving them to a 3D game, which is more challenging and time-consuming to develop. So, we may put thoughts to test in a 2D game before allocating a lot of resources to a 3D game which eventually leads to putting them in the reality.

\section{Background and Research Question}
It is important to not only establish trust but also maintain it between humans and robots in order to construct an efficient system. A trust violation occurs when one individual intentionally or unintentionally breaks the trust\cite{baker2018toward}. Competence and integrity are two important trustworthiness criteria used by researchers in the field of HHI \cite{mayer1995integrative}\cite{kim2009repair}\cite{sebo2019don}. The authors of the paper \cite{tolmeijer2020taxonomy} has created a taxonomy of trust-related failures and mitigation strategies. They classified the factors causing trust failure into four categories: design, system, user, and expectation. They have also differentiated the mitigation strategies which include apologies, explanations, and proposed alternatives by the system. \cite{esterwood2021you} have studied the effect on the trustworthiness after the apologies, denials, and promises made by a mechanical and anthropomorphic robot.
Though a lot of research has been done in the direction of trust violation strategies for various trust violations by the robot but not on how these trust violations affect the human-robot trust. Therefore in this research, we focus on the following research questions.\\
\textbf{Research Question 1:} Would two similar robot failures of equal scale affect human trust differently if one is caused by a moral-trust violation and the other by a performance-trust violation?\\
\textbf{Research Question 2:} Would change the mode of interaction from a grid-world game to a real-world 3D simulation influence human trust? \\
\\ 
We propose the following hypotheses: \\
\textbf{Hypothesis 1:} Given the 2D grid-world interface, where the robot's both moral and performance trust-violation have an equal magnitude in terms of individual and team score, moral-trust violation would have a different effect than performance-trust violation on human-robot trust. 
\\
\textbf{Hypothesis 2:} 
Changing the modality of the environment from a 2D grid-world to a 3D simulation will have equal or increased human-robot trust. 

\section{Methodology}
We have developed two games based on a search task in order to investigate our research questions. One is a website-hosted online 2D game \cite{khavas2021moral}. The other is an Unreal Engine-based 3D game. In both games, the player and a robotic agent will be determined to find hidden objects in the designated area. The game will be played for 10 rounds. We have established two types of scores to measure the performance of both robot and participants: individual score and team score. Individual score and team score are complementary to each other and so the robot or the participant can only maximize either one. In both games, we provide three distinct targets as shown in \ref{tab:t1}. Both the individual score and team score would be affected differently by each of the three targets. Target 1 will increase the team's score by 100 points while having no impact on individual performance. Targets 2 will reduce the team's score by 100 points while having no impact on individual scores. Target 3 will decrease the team score by 100 points while simultaneously increasing the individual score by 100 points. The participants will be assigned an agent who will assist them with the search task. 
The robot in this game exhibits three distinct behaviors:
\begin{enumerate}
    \item Robot violates the performance-trust: In this scenario, the robot will collect the target 2 which doesn't affect the individual score but decrease team score by 100 points. We can say that the robot's performance is not reliable in this situation.
    \item Robot violates the moral-trust: In this scenario,  will collect target 3 which increase the individual score by 100 points and decreases the team score by 100 points. We can say that the robot's behavior is non-ethical.
    \item Robot doesn't violate performance-trust or moral-trust: In this scenario, the robot will collect target 1 which doesn't affect individual score but increase team score by 100 points. We can say that the robot is encouraging the participants to increase the team score. 
\end{enumerate}

During the game, the participants will not be able to observe the agent's actions. After each round, the participants will be asked whether they wish to integrate or discard the agent's scores. The participants will make this judgment without knowing the robot's collected targets or the scores associated with them. Once they have made their decision, the robot's scores will be presented to the participants. Following the 10 rounds, the participants will be asked to complete a questionnaire related to the robot's performance and reliability.\par

Our study has been divided into two sections. In the first phase, we have focused on evaluating research question 1 where the participants will interact with a 2D grid world game to find the targets. During the second phase, we will be recruiting new participants for an in-person study where they will interact with a 3D interface to collect the targets. In the Section \ref{method2d} and \ref{method3d}, both the experiment designs are explained.
\begin{table}[h]
\begin{center}
\caption{Target types and effects of those on the scores}
\begin{tabular}{|p{1.3cm}|p{1.3cm}|p{1.3cm}|p{1.3cm}|p{1.3cm}|}
\hline \textbf{Targets}
  & \textbf{2D Environment} & \textbf{3D Environment} & \textbf{Team score} & \textbf{Individual score} \\ 
   \hline
 Target 1 & Gold Star & Fire Extinguisher & +100 & 0 \\
  \hline
Target 2 & Red Circle & Fire & -100 & 0 \\
 \hline
 Target 3 & Pink Triangle & Life potion & -100 & +100 \\
 \hline
\end{tabular}
\end{center}
\label{tab:t1} 
\end{table}
\subsection{2D environment} \label{method2d}
\begin{figure}[htbp]
\centerline{\includegraphics{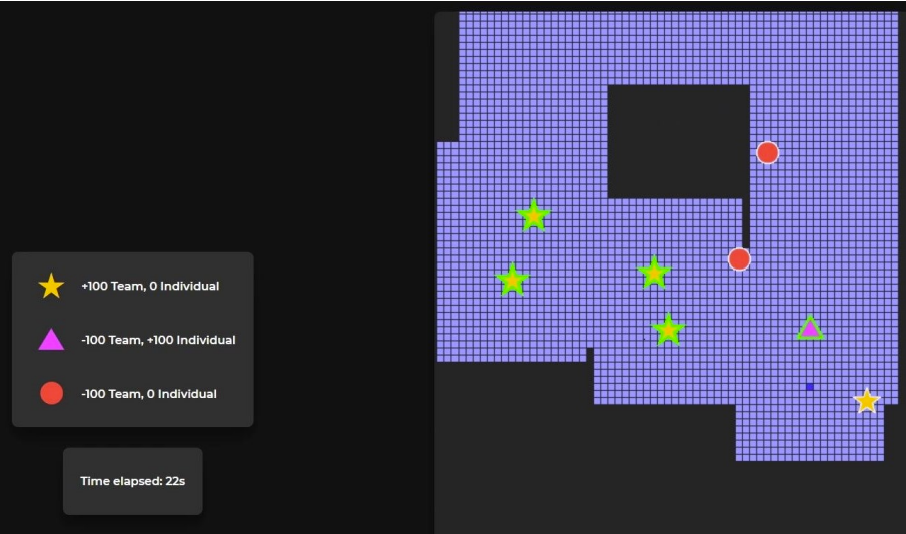}}
\caption{2D grid-world game interface\cite{khavas2021moral}}
\label{fig:2dgame1}
\end{figure}

In this environment, target 1 corresponds to gold star, target 2 corresponds to red circle and target 3 corresponds to pink triangle. Figure \ref{fig:2dgame1} shows an example of the 2D grid-world game interface where the human and robot have to search the area to find the targets and gain scores. As shown in Table I, the gold star increases team score by 100 points and red circle decrease the team score by 100 points and they both don't have any effect on the individual score. The pink triangle will decrease the team score by 100 points and increase the individual score by 100 points. We can say that the robot is portraying performance-trust violation by collecting red circles and moral-trust violation by collecting the pink triangles. We expect people to collect the red circle due to the robot's poor performance and the pink triangle due to the robot's lack of moral integrity. 

\subsection{3D environment} \label{method3d}
\begin{figure}[htbp]
\centerline{\includegraphics[width=8cm]{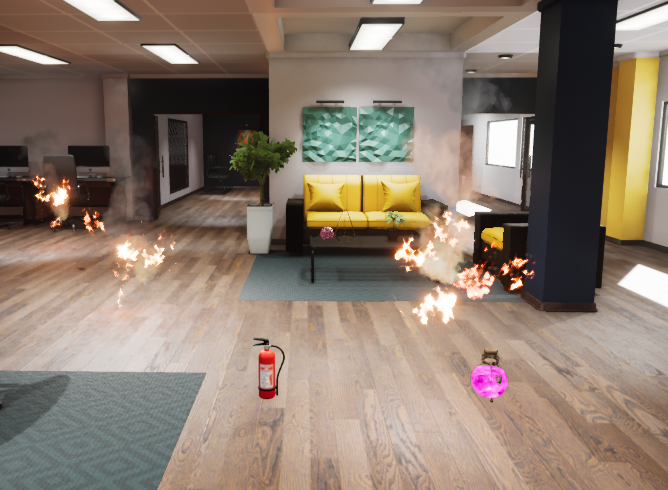}}
\caption{Example of 3D office environment}
\label{fig:office1}
\end{figure}

The user needs to have some realistic context of the task for the search and rescue mission. We observed from our prior experiment with the 2D game that the participants had to remember the appearance of all three targets and their corresponding individual and team scores. Therefore, we decided to built a realistic 3D environment using Unreal Engine.  It is an office environment that consists of modular tables, chairs, shelves, sofas, carpets, other furniture, office electronics, etc. Figure \ref{fig:office1} shows the developed 3D environment. In this environment, target 1 corresponds to a fire extinguisher, target 2 corresponds to fire and target 3 corresponds to a life potion. The fire will decrease the team score by 100 points and will not have any effect on the individual score. The fire extinguisher is a complement to fire which will increase the team score by 100 points but will not have any effect on the individual score. Lastly, the life potion will increase the individual score by 100 points and decrease the team score by 100 points.
The participants will be asked to explore around in the 3D environment using a mouse and four arrow keys on the keyboard. Once they get closer to the object, they can press the "e" key to collect the object. The main goal is to maximize either team or individual score by collecting the targets. The participant is also provided with the robot which is finding the objects simultaneously but the robot's actions are not shown to the participants. The robot's score will be shown once the participant decides whether to integrate or discard the robot's score. We can say that if the robot is collecting fire, the robot is not reliable as it is harming the team's performance but we can surely say that the robot's intention was not unethical as it did not gain any individual score by collecting fire. On the contrary, if the robot is collecting life potion, we can say that the robot is not ethical as it is gaining individual scores by compromising the team score. Table \ref{tab:t1} shows the various targets and their corresponding team and individual scores. 

We expect that the participants will have better intuition than the 2D game while picking up the objects. They will not have to remember the appearance of the objects as they can relate them to the real world. For example, it is natural to assume that a fire indicates danger and will have a negative consequence. A fire extinguisher, on the other hand, serves a useful purpose. 
Furthermore, playing a game in a 3D simulation will provide participants with a more engaging experience, allowing them to make more accurate and timely trust decisions.

\subsection{Participants and Trust Measurement}
In our first phase which was an online experiment, we hired around 100 participants to play the game with a grid world interface. 
It took 25-30 minutes for the participants to complete the experiment. The participants were paid \$4 after they complete the study\cite{khavas2021moral}.
For the second phase which is going to be an in-person study, we are planning to hire approximately 100 participants. They will be playing the simulation game for 10 rounds. 
It will take around 30 to 40 minutes for them to complete the experiment. They will be provided with \$20 gift card after they complete the study. \par
To analyze the impact of a robot's faulty behavior (violating performance trust and violating moral trust) on human-robot trust, we focus on the participants' responses collected at different time steps. In each round, once the participants complete the search task, they are asked whether they want to integrate the robot's score or not. Here, they will be making a blind decision as they are not aware of which targets did the robot collect in the current round. This process will decide the amount of trust between the robot and participants. After the participants make the trust decision, they will be provided with the information about the targets collected by the robot and the corresponding scores. This information will help them to conclude the robot's behavior and affect the trust decision for the next round. Moreover, in both 2D and 3D experiments, the participants will be filling out a Multi-Dimensional Measure of Trust (MDMT) \cite{ullman2019measuring} questionnaire with some additional questions about situation awareness for the 3D interface. Changing the modality of the environment from a 2D grid world to a 3D simulation will have equal or increased human-robot trust. 

\section{Conclusion and Future Work}

In this research, we aim to develop an interactive interface in order to study the trust relationship between humans and robots. We have designed two search games one for online study and another for in-person study to perform a between-subject study. In the online study, we aim to investigate the effect of robots violating moral-trust and performance-trust on humans. In the second phase, we try to investigate the same concept but use a more realistic interface. One of the main advantages of our proposed approach is that we can quickly and easily prototype different experiment designs in the 2D game before transferring them to a 3D game which is more complex and time-consuming to design. As a result, we can then test concepts before deploying them to a real-world environment.  \par
Currently, the participants are making a blind decision on whether to integrate or discard the robot's score meaning that they are not able to observe the robot's behavior while playing the game. In the future, we aim to introduce the robot entity in the 3D simulation which will portray various behaviors. For example not performing the search task efficiently, collecting only harmful objects, etc. The human and the robot will be searching the regions by collaborating with each other via real-time chatting where they can discuss the covered search area and collected targets. In this scenario, we would incorporate a new robot behavior which will be violating trust by misguiding the participants. \par
Another possible future direction would be elevating the trust measurement methods by incorporating physiological measurements such as electroencephalogram (EEG) signals and eye tracking. In addition to this, facial expression recognition can be used for the real-time assessment of trust. Once the robot has violated a trust whether it is performance-trust or moral-trust violation, it is necessary to study how the robot is going to build that trust again. Therefore, the robot could perform various gestures in order to regain the trust after it violates the trust. For example giving an explanation, making an apology, denying the violation, etc\cite{de2018automation}. It will be interesting to study the trust pattern after the robot performs trust repair strategies. 

\bibliographystyle{plain}
\bibliography{ref.bib}
\end{document}